\documentclass[letterpaper]{article} 
\usepackage{aaai25}  
\usepackage{times}  
\usepackage{helvet}  
\usepackage{courier}  
\usepackage[hyphens]{url}  
\usepackage{graphicx} 
\usepackage{natbib}  
\usepackage{caption} 
\usepackage{algorithm}
\usepackage{algorithmic}

\usepackage{newfloat}
\usepackage{listings}
\DeclareCaptionStyle{ruled}{labelfont=normalfont,labelsep=colon,strut=off} 
\floatstyle{ruled}
\newfloat{listing}{tb}{lst}{}
\floatname{listing}{Listing}
\usepackage{dsfont}
\newcommand\dsone{\mathds{1}}
\usepackage{float}
\usepackage{multirow}
\usepackage{subcaption}
\usepackage{booktabs} 
\usepackage{adjustbox}
\title{AM Flow: Adapters for Temporal Processing in Action Recognition}
\author{
        Tanay Agrawal\equalcontrib,
    Abid Ali\equalcontrib,
     Antitza Dantcheva,
    Francois Bremond}
\affiliations{
    INRIA, France\\

}

\usepackage{bibentry}

\usepackage{amssymb}

\begin{document}

\maketitle

\begin{abstract}
Deep learning models, in particular \textit{image} models, have recently gained generalisability and robustness. 
In this work, we propose to exploit such advances in the realm of \textit{video} classification. 
Video foundation models suffer from the requirement of extensive pretraining and a large training time. 
Towards mitigating such limitations, we propose "\textit{Attention Map (AM) Flow}" for image models, a method for identifying pixels relevant to motion in each input video frame. In this context, we propose two methods to compute AM flow, depending on camera motion.
AM flow allows the separation of spatial and temporal processing, while providing improved results over combined spatio-temporal processing (as in video models). 
Adapters, one of the popular techniques in parameter efficient transfer learning, facilitate the incorporation of AM flow into pretrained image models, mitigating the need for full-finetuning. 
We extend adapters to "\textit{temporal processing adapters}" by incorporating a temporal processing unit into the adapters.
Our work achieves faster convergence, therefore reducing the number of epochs needed for training.
Moreover, we endow an image model with the ability to achieve state-of-the-art results on popular action recognition datasets. This reduces training time and simplifies pretraining. 
We present experiments on Kinetics-400, Something-Something v2, and Toyota Smarthome datasets, showcasing state-of-the-art or comparable results. 
Our code will be made available on github.
\end{abstract}

%

\section{Introduction}
\label{sec1}
Currently, foundation models have advanced in multiple domains, on the task, for which they were explicitly trained, as well as on other implicit applications. For example, \cite{karamcheti2023languagedriven} show that LLMs work well as pattern prediction machines, in addition to language-related tasks. The hypothesis behind this work is inspired by this phenomenon. 
While image models do not comprise motion, attention maps computed in transformer blocks are endowed with the ability to derive pixels which are pertinent to motion.

\begin{figure}[t]
\begin{center}
   \includegraphics[width=0.9\linewidth]{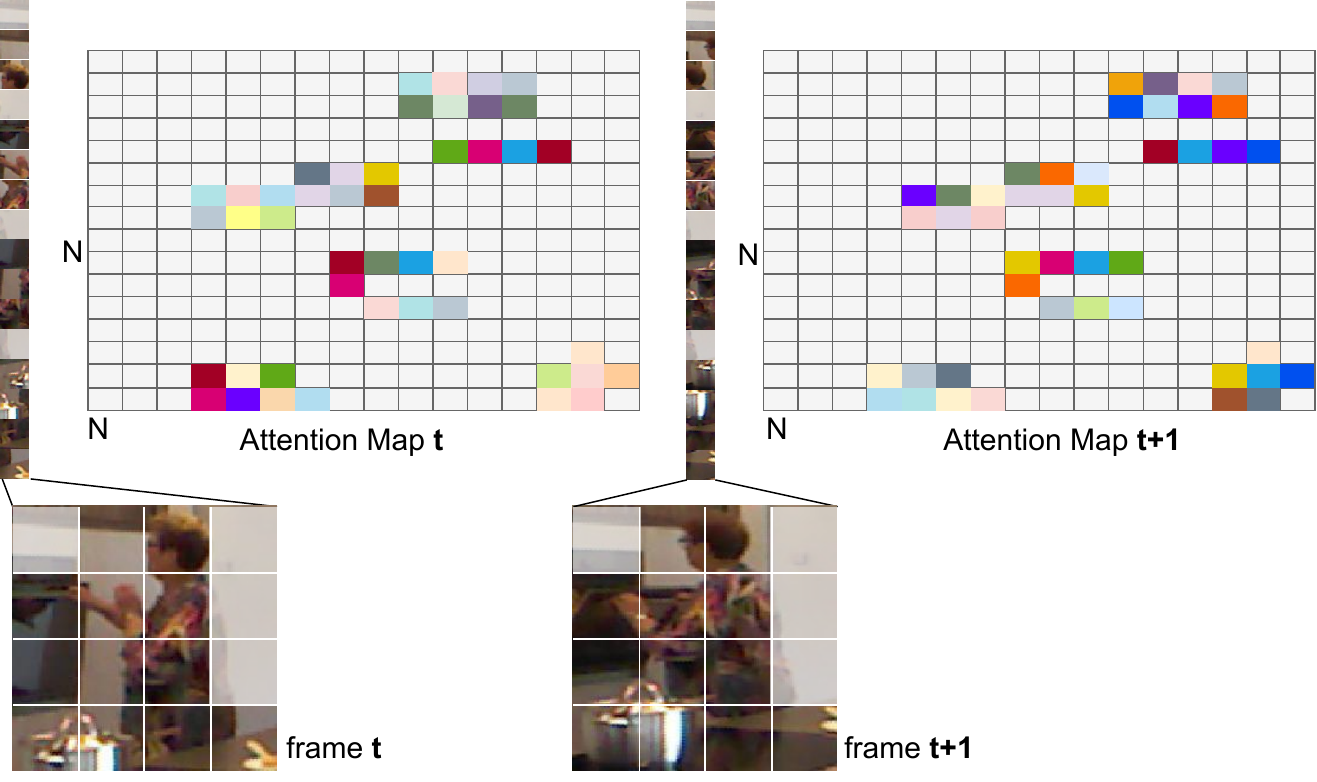}
   \caption{\textbf{Intuition of AM flow}. No motion in the videos (frames) and minimal change in the attention map are represented by white/transparent. Here, rows with changes in the attention map correspond to input patches with motion, for example second and third patches have motion and the second and third rows of the attention maps have change. \textit{Note: the attention map is hand-crafted only for explanation and the colours have no intended meaning.}}
\label{fig1}
\end{center}
\end{figure} 

Despite their remarkable capabilities, video foundation models come with significant limitations pertaining to their resource-intensive nature. Specifically, such models necessitate a substantial amount of pretraining data and training time, posing practical constraints. 
This work addresses this limitation using an image foundation model as a backbone along with the proposed modules: AM Flow and temporal processing adapters - an alternative to video processing backbones, increasing performance and reducing training time compared to the state-of-the-art (SOTA).

Image foundation models have seen massive growth with works such as Dinov2 \cite{oquab2023dinov2} and Hiera \cite{ryali2023hiera}. This growth is due to improved spatial attention, which can be exploited to alleviate the above challenge with video models. Instead of learning fine-grained temporal relations directly from videos, the \textit{absolute difference} of attention maps (taken from transformer encoders) for two consecutive frames provides simplified, encoded information about the motion in the frames. Here, we refer to this difference as \textbf{A}ttention \textbf{M}ap flow or \textbf{AM} flow. Figure \ref{fig1} illustrates the intuition behind computing AM flow. This rich information can be achieved with only an absolute difference operation performed on the attention maps, and therefore resulting in a reduced training and inference time.

We aim to remove temporal processing from the backbone, in order to expedite convergence and let temporal processing adapters account for time. 
It is counter-intuitive that separating spatial and temporal processing improves performance. However, end-to-end training allows the network to learn the relationship between space and time through backpropagation. 
Adapters are a parameter-efficient transfer learning technique, in which modules are added to a frozen pre-trained backbone and only these modules are trained during the new transfer learning protocol. They work by adapting the distribution of the original network to the new task, data, or even a new modality.


Therefore, by combining temporal processing adapters and AM flow, we propose an expedite computation of spatio-temporal relations for video classification. 
AM flow can simply be concatenated with the input to the adapters. 
Section \emph{"Methodology"} discusses this in detail. A benefit of our method has to do with the downsampled embedding from this adapter, providing rich spatial information, pertaining to motion. 
We conduct experiments with three temporal processing modules, namely transformer encoder, TCN, and LSTM. These are added to the temporal processing adapters and our experiments suggest that all perform well.


To summarise, the contributions of our work include the following.
\begin{itemize}
    \item We introduce AM flow, an efficient method to compute motion between consecutive frames.
    \item We propose two methods to compute AM flow, depending on camera motion.
    \item We propose a novel architecture to incorporate AM flow using temporal processing adapters.
    \item We achieve SOTA performance on three datasets, namely Kinetics-400, Something-something v2, and Toyota Smarthome.
\end{itemize}


\begin{figure*}[ht]
\begin{center}
   \includegraphics[width=0.85\linewidth]{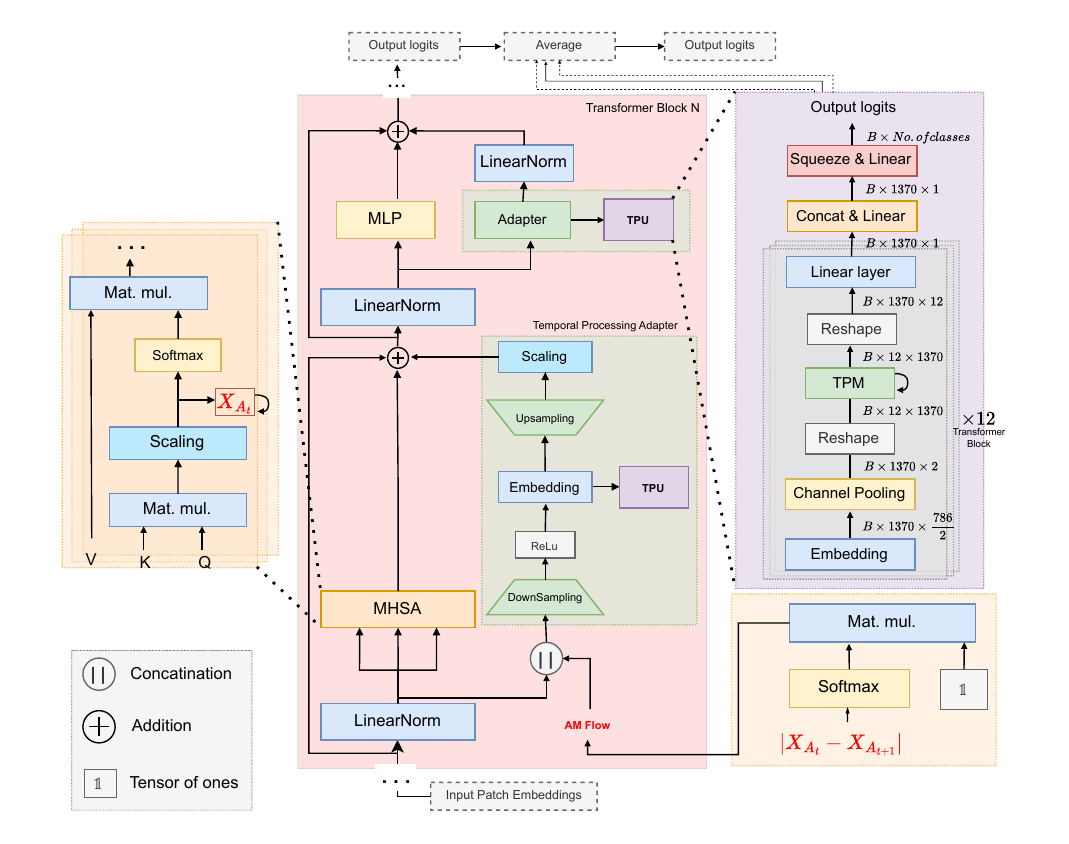}
   \caption{The middle part of the figure shows 
   the frozen image model (ViT) (in red colour) with trainable additions: temporal processing adapters (in green) containing temporal processing units (in purple). The adapter across MHSA takes AM flow as input along with the input to the transformer block. $X_{A_t}$ is shown as computed inside $MHSA$ on the left (in yellow). On the right, $X_{A_t}$ and $X_{A_{t+1}}$ are used to compute $AM flow$ (in yellow). $t$ and $t+1$ signify different time-steps for the input frames. (Violet) shows the global temporal processing unit (TPU) and the classification head (squeeze and linear) added to it. All logits received from the temporal processing module (TPM) and the frozen model branch are averaged to obtain the final classification logits.} 
   \label{fig2}
\end{center}
\end{figure*}

\section{Related Work}
\label{sec2}

\subsection{Video Classification}
Deep learning algorithms have mainly dominated action recognition in unconstrained videos, owing to the availability of huge video datasets, such as Kinetics \cite{carreira2017quo} and Something-Something \cite{goyal2017something}. The model architectures used by the existing video algorithms have evolved from CNNs \shortcite{karpathy2014large,feichtenhofer2020x3d, feichtenhofer2019slowfast,lin2019tsm, liu2021tam, pan2021actor,tran2015learning,xie2018rethinking, wang2018temporal} to Transformers \shortcite{arnab2021vivit, fan2021multiscale, vaswani2017attention, li2023uniformer, li2022mvitv2}. Since the Vision Transformer (ViT)\cite{dosovitskiy2020image} has emerged as a new paradigm in computer vision, video understanding transformers have been intensively researched by extending pretrained image models. Pre-trained image transformers were utilised to initialise the component of the video transformers \shortcite{arnab2021vivit, bertasius2021space, yan2022multiview, dosovitskiy2020image} or to inflate the video transformers \cite{liu2022video}. Although transformers have shown improved performance in video action recognition, they require extensive finetuning on video datasets, making training inefficient. Our method can achieve the same performance with approximately 10 times fewer training epochs compared to fully finetuning.

Utilising intermediate features for classification has also been explored previously. \citeauthor{delta_distillation,TSM} explore this along with other works using motion related intermediate features \cite{representation_flow}. However, none of them explored the rich relational information present in the attention map of transformer blocks, and AM Flow is defined to exploit this oversight and allow this work to achieve SOTA results.

\subsection{Parameter Efficient Transfer Learning}
The task of efficient tuning has attracted increased interest in NLP as a result of the wider deployment of large pretrained language models across a variety of downstream tasks \shortcite{houlsby2019parameter, chen2022adaptformer, jie2022convolutional, nie2022protuning, sung2022vladapter}. In recent years, parameter-efficient fine-tuning techniques, such as adapters, have been widely used in computer vision.
\shortcite{stadapter, dualpath, chen2022adaptformer, ju2022prompting, lin2022frozen, ni2022expanding, sung2022vladapter}. These works demonstrated that pretrained image models can be good video learners. \cite{li2022mvitv2} learnt an additional decoder with 3D convolution layers and cross-frame attention, and \cite{liu2022video} placed additional depth-wise 3D convolution layers between the adapter's down/up-projection layers to do temporal reasoning, causing computational inefficiencies. \citeauthor{jia2022visual, liu2022swin} necessitate the use of an additional text-encoder branch as a classifier. 
Most recently, the ST-adapter \cite{stadapter} proposed adding depthwise convolution to the input to pool the temporal dimension, which reduces the model's capability to process time. \citeauthor{dualpath} use adapters in two separate pathways for spatial and temporal learning, but they are limited by the number of frames they can input as they stack frames in a grid in one frame to represent time. Our work overcomes these two limitations, as we do not pool time in the early stages of the network and are not limited by design in the number of frames we can input to the model.

\section{Methodology}
\label{sec3}
In this section, we walk the reader through the background for each contribution and the details of the final architecture used for experiments. Figure \ref{fig2} illustrates the architecture of the model. We take ViT trained using Dinov2 \cite{oquab2023dinov2} as our image model and freeze its weights. Then, scaled parallel adapters are added to the model, and AM flow is concatenated with adapters across the self-attention modules. Various temporal processing modules (TPMs) along with the image model branch give output logits which are averaged, giving the final output. Details are given below.

\subsection{Foundation Models and Adding Adapters}
Today, most of the foundation models in computer vision are transformer-based. This work is exclusively focused on these. 
As stated above, we choose ViT trained using Dinov2 for this work. 
Adapters are added to individual transformer encoders/blocks, making this method applicable to all model architectures. 
For completeness, we will start with a background for our choice of adapters.

As discussed in the previous section, there are many variations of adapters. Most of them use the basic block of adapters and change how it is used according to need, where they are placed, and also the architecture of the block itself. This basic block is:
\begin{math}
Adapter(x) = x + f(xW_{down})W_{up}
\end{math}. $f$ is usually $ReLU$. We choose our adapters based on the functions we want them to serve.
We have two requirements from an adapter: 
\begin{itemize}
    \item Adapting to the spatial distribution of the dataset to be trained on.
    \item Providing a downsampled embedding to be used as input to the temporal processing module.
\end{itemize}

\subsubsection{Adapting to a new spatial distribution}
\begin{figure}[ht]
\begin{center}
   \includegraphics[width=0.5\linewidth]{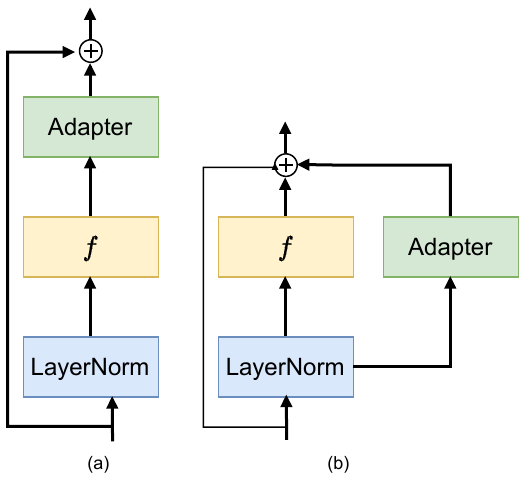}
   \caption{(a) Serial Adapters. (b) Parallel Adapters} \label{fig3}
\end{center}
\end{figure}

\cite{dualpath} uses the basic adapter block in two settings: serial and parallel. They show that parallel adapters work well for spatial adaptation and serial for temporal adaptation. The two settings are shown in Figure \ref{fig3}.

Following experiments, we concur that for our purpose parallel adapters work better. Specifically, we use scaled parallel adapters \cite{he2022towards}, which is a better choice over parallel adapters. They can be written as:
\begin{equation}
Adapter(x) = x + s.f(xW_{down})W_{up}
\end{equation} Here, $s$ is a learnable scalar and $f$ is the activation function.

\subsubsection{Embedding for temporal reasoning}
\label{3.1.2}
The basic block of adapters works well for this purpose, as they inherently provide a downsampled embedding. Parallel adapters only need to learn the change in distribution to be added to the pretrained model. On the other hand, serial adapters accommodate this change along with the information that passes through the model. Therefore, for our requirements, serial adapters are traditionally believed to work better as the information in the bottleneck layer is more rich. However, the addition of AM flow forces parallel adapters to learn substantial information passing through the network too, this is discussed in Section \emph{Additional Experiments and Discussion}. This additional information bridges the gap, allowing parallel adapters to work as well as serial.

\subsection{Adding AM Flow and Modifying Attention}
Building upon the previous step, we have a frozen image model with adapters added to it for finetuning, but, we are missing the benefit of spatio-temporal attention in video models. To answer this, we introduce the AM flow. 
It examines two temporal frames to detect local movement and informs the adapters so that they can account for it when adjusting the spatial attention in the image model.
Thus, we achieve spatio-temporal attention with only a minuscule number of parameters added to the original model.

To compute AM flow, we use the attention maps from a transformer block. An attention map corresponding to the frame $t$ ($X_{A_t} \in \mathbb{R}^{N\times N}$) is a matrix of size $N\times N$, where N is the number of embeddings in the input to the transformer block. For us (and also commonly in most transformer-based architectures), $N$ is equal to the number of patches of the input. So, the attention map is a matrix in which each entry signifies the relationship of two particular patches of the input. Taking the absolute difference of this matrix for two consecutive inputs results in a matrix where each entry signifies the amount of change in the relationship of the two particular patches. Taking a row-wise sum of the transpose of this matrix, we get the amount of change in each input patch. The softmax and addition are performed along different dimensions of the attention map. The addition aggregates the importance of a particular pixel with respect to all the others and softmax normalises it. So, the difference of normalised, aggregated significance of pixels for two frames gives AM Flow. Mathematically, this can be expressed as:

\begin{equation}
    X_{A_t} = \frac{Q_tK_t^T}{\sqrt{d_K}}
\end{equation} As shown in Figure \ref{fig2} (in yellow), $X_{A_t}$ is an intermediate term used to compute AM flow. The rest of the symbols have the common meaning of self-attention.

\begin{equation}
    AM flow = softmax(|X_{A_t} - X_{A_{t+1}}|)^T \cdot \dsone
\end{equation} 
$\dsone \in \mathbb{R}^{N\times 1}$ is a matrix of the same shape as $Q, K, V$, but filled with ones. The multiplication operation is equivalent to the row-wise sum of a matrix. We use $\dsone$ in place of $V$ (which is commonly used in attention) as AM flow needs to highlight the position of the relevant patches and contextual information provided by the input is not required by it.

\subsubsection{Handling irrelevant motion - Aligning Encoder}
\begin{figure}[ht]
\begin{center}
   \includegraphics[width=0.45\linewidth]{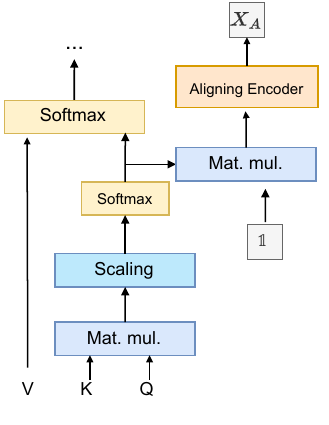}
   \caption{This figure shows how AM flow ($X_A$) is computed in case there is a camera movement or motion in the background} \label{fig4}
\end{center}
\end{figure}

AM flow as described above is based on the assumption: the camera is stable. In the case of moving cameras, the computed flow is very noisy unless the spatial difference between two consecutive frames is reduced. This can be solved in preprocessing, for example, by taking a crop of the subject performing the action using foreground segmentation. Otherwise, this would have to be corrected by the model. 
Therefore, before computing the absolute difference, we take the row-wise sum of the attention maps and pass it through a transformer encoder (called Aligning Encoder). \footnote{We use fast attention\cite{fast_attention} for the aligning encoder.}
The transformer encoder learns to align the relevant information in attention maps and, in turn, disregards the rest. It still has information about global motion owing to backpropogation from the temporal processing module. The weights of the aligning encoder are shared between the two frames used to compute AM flow. Figure \ref{fig4} illustrates the aligning encoder added to self-attention in a transformer block of the frozen image model. Mathematically expressed as:

\begin{equation}
    X_{A_t} = Aligning Encoder(softmax(\frac{Q_tK_t^T}{\sqrt{d_K}})^T \cdot \dsone)
\end{equation} The variables have the same references as above. The aligning encoder is a simple transformer encoder with 12 heads.

\begin{equation}
    AM Flow = |X_{A_t} - X_{A_{t+1}}|
\end{equation}

\subsection{Classification Head and Temporal Processing}
With the above step, we have successfully merged local temporal attention with spatial attention in a frame. To extend this to video classification, we need global temporal processing. It is very interesting to note the robustness of the rich downsampled embedding from the adapter. It allows freedom in the choice of the TPM. Figure \ref{fig2} (in violet) demonstrates the temporal and classification module (TPU) and is self-explanatory. The formulation of the input embedding can be expressed as:

\begin{equation}
    E = ReLU(W_{down}(Layer Norm(X) || AM Flow))
\end{equation} $E$ is the embedding and $W_{down}$ is the downsampling layer of the adapter.

Thus, we are able to obtain SOTA results with various TPMs. 

The final step of classification is managing the 25 outputs we get (for ViT-L). 2 from each transformer block as there are 2 adapters in each block and 1 from the frozen image model taking the final frame as input. There are 12 blocks in ViT-B. We initially added a linear voting layer, but it just makes the training unstable. Average pooling works well here. An interesting thing to note is that even a randomly initialised frozen linear layer works well in place of average pooling. The scaling parameter in the adapters learns to work around the randomly assigned weights.

\section{Experiments and Observations} \label{sec4}

\subsection{Datasets}
We evaluate our model on three popular datasets with standard preprocessing. Something-something v2 \cite{goyal2017something} is a dataset with fixed cameras having different angles and multiple objects of interest. Kinetics-400 \cite{carreira2017quo} is a dataset with moving cameras, and there is a single person of interest. Toyota Smarthome \cite{das2019toyota} has a fixed camera and one target person. The details of the datasets are as follows:

\textbf{Something-Something v2}
 (SSv2) is a more challenging dataset, since it requires strong temporal modelling. It contains about 168.9K training videos and 24.7K validation videos for 174 classes.
 \textbf{Kinetics-400}
 (K400) is a large dataset. It contains about 240K training videos and 20K validation videos for 400 human action categories. Each video is trimmed to have a length of 10 seconds. Although the K400 dataset provides a wide range of categories, they are known to be highly biased in spatial appearance.
 \textbf{Toyota Smarthome}
 (Smarthome) is a small dataset. It contains 16.1k video clips and 31 complex daily-life activities performed naturally without strong prior instructions. For the evaluation of this dataset, we follow the cross-subject (CS) protocol proposed in \cite{das2019toyota}.

\begin{table*}[h]
\centering
\resizebox{0.9\linewidth}{!}{%
\begin{tabular}{lccccccc}
\hline
\multicolumn{1}{c|}{}                                  & \multicolumn{1}{c|}{}                                    & \multicolumn{1}{c|}{}                               &                                                                                             & \multicolumn{1}{c|}{}                                                                                                & \multicolumn{1}{c|}{}                                                                               & \multicolumn{2}{c}{\textbf{SSV2}}                             \\
\multicolumn{1}{c|}{\multirow{-2}{*}{\textbf{Method}}} & \multicolumn{1}{c|}{\multirow{-2}{*}{\textbf{Pretrain}}} & \multicolumn{1}{c|}{\multirow{-2}{*}{\textbf{\#F}}} & \multirow{-2}{*}{\textbf{\begin{tabular}[c]{@{}c@{}}Model\\ \# Params \\ (M)\end{tabular}}} & \multicolumn{1}{c|}{\multirow{-2}{*}{\textbf{\begin{tabular}[c]{@{}c@{}}Trainable\\ \# Params \\ (M)\end{tabular}}}} & \multicolumn{1}{c|}{\multirow{-2}{*}{\textbf{\begin{tabular}[c]{@{}c@{}}FLOPs\\ (T)\end{tabular}}}} & \textbf{Top-1}                & \textbf{Top-5}                \\ \hline
\multicolumn{8}{l}{{\textit{Specialised backbone with supervised pretraining}}}                                                                                                                                                                                                                                                                                                                                                                                                                                                               \\
\multicolumn{1}{l|}{MViTv1-B \cite{fan2021multiscale}}                  & \multicolumn{1}{c|}{K400}                                & \multicolumn{1}{c|}{32}                             & \multicolumn{1}{c|}{37}                                                                     & \multicolumn{1}{c|}{37}                                                                                              & \multicolumn{1}{c|}{4.1}                                                                            & 67.7                          & 90.9                          \\
\multicolumn{1}{l|}{VideoSwin-B \cite{liu2022videoswin}}                       & \multicolumn{1}{c|}{IN-21K+K400}                         & \multicolumn{1}{c|}{32}                             & \multicolumn{1}{c|}{89}                                                                     & \multicolumn{1}{c|}{89}                                                                                              & \multicolumn{1}{c|}{1.0}                                                                            & 69.6                          & 92.7                          \\
\multicolumn{1}{l|}{MViTv2-L \cite{li2022mvitv2}}                          & \multicolumn{1}{c|}{IN-21K+K400}                         & \multicolumn{1}{c|}{40}                             & \multicolumn{1}{c|}{213}                                                                    & \multicolumn{1}{c|}{213}                                                                                             & \multicolumn{1}{c|}{8.5}                                                                            & 73.3                          & 92.7                          \\ \hline
\multicolumn{8}{l}{\textit{Vanilla ViT with self-supervised pretraining for 1600 epochs.}}                                                                                                                                                                                                                                                                                                                                                                                                                                                                         \\
\multicolumn{1}{l|}{VideoMAE-B \cite{tong2022videomae}}                        & \multicolumn{1}{c|}{-}                                   & \multicolumn{1}{c|}{16}                             & \multicolumn{1}{c|}{87}                                                                     & \multicolumn{1}{c|}{87}                                                                                              & \multicolumn{1}{c|}{1.1}                                                                            & 70.8                          & 92.4                          \\
\multicolumn{1}{l|}{VideoMAE-L \cite{tong2022videomae}}                        & \multicolumn{1}{c|}{-}                                   & \multicolumn{1}{c|}{16}                             & \multicolumn{1}{c|}{305}                                                                    & \multicolumn{1}{c|}{305}                                                                                             & \multicolumn{1}{c|}{3.6}                                                                            & 74.3                          & 94.6                          \\ \hline
\textit{Well-prepared ViT with plug-and-play modules.} & \multicolumn{1}{l}{\textit{}}                            & \multicolumn{1}{l}{\textit{}}                       & \multicolumn{1}{l}{\textit{}}                                                               & \multicolumn{1}{l}{\textit{}}                                                                                        & \multicolumn{1}{l}{\textit{}}                                                                       & \multicolumn{1}{l}{\textit{}} & \multicolumn{1}{l}{\textit{}} \\
\multicolumn{1}{l|}{TimeSformer-L \cite{bertasius2021space}}                     & \multicolumn{1}{c|}{IN-21K}                              & \multicolumn{1}{c|}{96}                             & \multicolumn{1}{c|}{121}                                                                    & \multicolumn{1}{c|}{121}                                                                                             & \multicolumn{1}{c|}{7.1}                                                                            & 62.3                          & 81.0                          \\
\multicolumn{1}{l|}{MTV-B \cite{yan2022multiview}}                             & \multicolumn{1}{c|}{IN-21K+K400}                         & \multicolumn{1}{c|}{32}                             & \multicolumn{1}{c|}{310}                                                                    & \multicolumn{1}{c|}{310}                                                                                             & \multicolumn{1}{c|}{11.2}                                                                           & 68.5                          & 90.4                          \\
\multicolumn{1}{l|}{CoVeR \cite{zhang2021cotraining}}                             & \multicolumn{1}{c|}{JFT-3B+KMI}                          & \multicolumn{1}{c|}{16}                             & \multicolumn{1}{c|}{431}                                                                    & \multicolumn{1}{c|}{431}                                                                                             & \multicolumn{1}{c|}{17.6}                                                                           & 70.8                          & -                             \\ \hline
\multicolumn{8}{l}{\textit{Full tuning}}                                                                                                                                                                                                                                                                                                                                                                                                                                                                                                                           \\
\multicolumn{1}{l|}{UniFormerV2-B \cite{li2023uniformerv2}}                     & \multicolumn{1}{c|}{CLIP-400M}                           & \multicolumn{1}{c|}{32}                             & \multicolumn{1}{c|}{163}                                                                    & \multicolumn{1}{c|}{163}                                                                                             & \multicolumn{1}{c|}{1.1}                                                                            & 69.5                          & 92.3                          \\
\multicolumn{1}{l|}{UniFormerV2-L \cite{li2023uniformerv2}}                     & \multicolumn{1}{c|}{CLIP-400M}                           & \multicolumn{1}{c|}{32}                             & \multicolumn{1}{c|}{574}                                                                    & \multicolumn{1}{c|}{574}                                                                                             & \multicolumn{1}{c|}{5.2}                                                                            & 73.0                          & 94.5                          \\
\multicolumn{1}{l|}{UniFormerV2-B \cite{li2023uniformerv2}}                     & \multicolumn{1}{c|}{CLIP-400M}                           & \multicolumn{1}{c|}{16}                             & \multicolumn{1}{c|}{163}                                                                    & \multicolumn{1}{c|}{163}                                                                                             & \multicolumn{1}{c|}{0.6}                                                                            & 69.5                          & 92.3                          \\
\multicolumn{1}{l|}{UniFormerV2-L \cite{li2023uniformerv2}}                     & \multicolumn{1}{c|}{CLIP-400M}                           & \multicolumn{1}{c|}{16}                             & \multicolumn{1}{c|}{574}                                                                    & \multicolumn{1}{c|}{574}                                                                                             & \multicolumn{1}{c|}{8.0}                                                                            & 72.1                          & 93.6                          \\ \hline
\multicolumn{8}{l}{\textit{Parameter Efficient Tuning}}                                                                                                                                                                                                                                                                                                                                                                                                                                                                                                            \\
\multicolumn{1}{l|}{AdaptFormer-B \cite{chen2022adaptformer}}                     & \multicolumn{1}{c|}{CLIP-400M}                           & \multicolumn{1}{c|}{8}                              & \multicolumn{1}{c|}{94}                                                                     & \multicolumn{1}{c|}{8}                                                                                               & \multicolumn{1}{c|}{0.5}                                                                            & 51.3                          & 70.6                          \\
\multicolumn{1}{l|}{Pro-tuning-B \cite{nie2022protuning}}                      & \multicolumn{1}{c|}{CLIP-400M}                           & \multicolumn{1}{c|}{95}                             & \multicolumn{1}{c|}{9}                                                                      & \multicolumn{1}{c|}{8}                                                                                               & \multicolumn{1}{c|}{9.6}                                                                            & 66.7                          & -                             \\
\multicolumn{1}{l|}{ST-Adapter \cite{stadapter}}                        & \multicolumn{1}{c|}{CLIP-400M}                           & \multicolumn{1}{c|}{32}                             & \multicolumn{1}{c|}{97}                                                                     & \multicolumn{1}{c|}{11}                                                                                              & \multicolumn{1}{c|}{2}                                                                              & 69.5                          & 92.6                          \\
\multicolumn{1}{l|}{DUALPATH-B \cite{dualpath}}                        & \multicolumn{1}{c|}{CLIP-400M}                           & \multicolumn{1}{c|}{16}                             & \multicolumn{1}{c|}{99}                                                                     & \multicolumn{1}{c|}{13}                                                                                              & \multicolumn{1}{c|}{0.7}                                                                            & 70.3                          & 92.9                          \\
\multicolumn{1}{l|}{DUALPATH-L \cite{dualpath}}                        & \multicolumn{1}{c|}{CLIP-400M}                           & \multicolumn{1}{c|}{48}                             & \multicolumn{1}{c|}{336}                                                                    & \multicolumn{1}{c|}{33}                                                                                              & \multicolumn{1}{c|}{1.9}                                                                            & 72.2                          & 93.7                          \\ \hline
\multicolumn{1}{l|}{\textbf{Ours - AM/12, TCN (Dinov2-B)}}                     & \multicolumn{1}{c|}{\textbf{IN-21K}}                   & \multicolumn{1}{c|}{\textbf{24}}                     & \multicolumn{1}{c|}{\textbf{\textbf{86+28+45}+54}}                                                              & \multicolumn{1}{c|}{\textbf{28+45+54}}                                                                                       & \multicolumn{1}{c|}{\textbf{5.1}}                                                                      & \textbf{74.8}                     & \textbf{95.0}                     \\
\multicolumn{1}{l|}{Ours - AM/12, TCN (CLIP-B)}                     & \multicolumn{1}{c|}{IN-21K}                   & \multicolumn{1}{c|}{24}                     & \multicolumn{1}{c|}{86+28+45+54}                                                              & \multicolumn{1}{c|}{28+45+54}                                                                                       & \multicolumn{1}{c|}{5.1}                                                                      & {73.5}                     & {94.7}                     \\
\multicolumn{1}{l|}{{Ours - AM/12, Transformer (Dinov2-B)}}                     & \multicolumn{1}{c|}{{IN-21K}}                   & \multicolumn{1}{c|}{{24}}                     & \multicolumn{1}{c|}{{{86+28+45}+103}}                                                              & \multicolumn{1}{c|}{{28+45+103}}                                                                                       & \multicolumn{1}{c|}{{11.9}}                                                                      & {74.6}                     & {95.0}                     \\
\multicolumn{1}{l|}{Ours - AM/12, LSTM (Dinov2-B)}                        & \multicolumn{1}{c|}{IN-21K}                           & \multicolumn{1}{c|}{8}                             & \multicolumn{1}{c|}{86+30+45+360}                                                                    & \multicolumn{1}{c|}{30+45+360}                                                                                              & \multicolumn{1}{c|}{4.6}                                                                            &        58.3                  &               82.8             \\
\hline
\end{tabular}
}
\caption{Comparison with the SOTA on SSv2. The parameters are broken into components: backbone, {AM flow and linear layers}, {aligning encoder}, and {temporal processing module}. AM/12 signifies that AM flow is added to all 12 transformer blocks.}
\label{ssv2}
\end{table*}

\begin{table*}[h]
\centering
\resizebox{\linewidth}{!}{%
\begin{tabular}{lcccccccc}
\hline
\multicolumn{1}{c|}{}                                  & \multicolumn{1}{c|}{}                                    & \multicolumn{1}{c|}{}                                    & \multicolumn{1}{c|}{}                                 &                                                                                             & \multicolumn{1}{c|}{}                                                                                                & \multicolumn{1}{c|}{}                                                                               & \multicolumn{2}{c}{\textbf{K400}} \\
\multicolumn{1}{c|}{\multirow{-2}{*}{\textbf{Method}}} & \multicolumn{1}{c|}{\multirow{-2}{*}{\textbf{Backbone}}} & \multicolumn{1}{c|}{\multirow{-2}{*}{\textbf{Pretrain}}} & \multicolumn{1}{c|}{\multirow{-2}{*}{\textbf{Views}}} & \multirow{-2}{*}{\textbf{\begin{tabular}[c]{@{}c@{}}Model\\ \# Params \\ (M)\end{tabular}}} & \multicolumn{1}{c|}{\multirow{-2}{*}{\textbf{\begin{tabular}[c]{@{}c@{}}Trainable\\ \# Params \\ (M)\end{tabular}}}} & \multicolumn{1}{c|}{\multirow{-2}{*}{\textbf{\begin{tabular}[c]{@{}c@{}}FLOPs\\ (T)\end{tabular}}}} & \textbf{Top-1}  & \textbf{Top-5}  \\ \hline
\multicolumn{9}{l}{\textit{Specialised backbone with supervised pretraining}}                                                                                                                                                                                                                                                                                                                                                                                                                                                                                                \\
\multicolumn{1}{l|}{MViTv1-B \cite{fan2021multiscale}}                 & \multicolumn{1}{c|}{MViTv1-B}                            & \multicolumn{1}{c|}{}                                    & \multicolumn{1}{c|}{64 $\times$ 3 $\times$ 3}         & \multicolumn{1}{c|}{37}                                                                     & \multicolumn{1}{c|}{37}                                                                                              & \multicolumn{1}{c|}{4.1}                                                                            & 81.2            & 95.1            \\
\multicolumn{1}{l|}{VideoSwin-L \cite{liu2022videoswin}}                       & \multicolumn{1}{c|}{Swin-L}                              & \multicolumn{1}{c|}{IN-21K}                              & \multicolumn{1}{c|}{32 $\times$ 3 $\times$ 4}         & \multicolumn{1}{c|}{197}                                                                    & \multicolumn{1}{c|}{197}                                                                                             & \multicolumn{1}{c|}{7.2}                                                                            & 83.1            & 95.9            \\
\multicolumn{1}{l|}{MViT-L \cite{li2022mvitv2}}                            & \multicolumn{1}{c|}{MViTv2-L}                            & \multicolumn{1}{c|}{IN-21K}                              & \multicolumn{1}{c|}{40 $\times$ 3 $\times$ 5}         & \multicolumn{1}{c|}{218}                                                                    & \multicolumn{1}{c|}{218}                                                                                             & \multicolumn{1}{c|}{42.4}                                                                           & 86.1            & 97.0            \\ \hline
\multicolumn{9}{l}{\textit{Vanilla ViT with self-supervised pretraining for 1600 epochs.}}                                                                                                                                                                                                                                                                                                                                                                                                                                                                                                          \\
\multicolumn{1}{l|}{VideoMAE-B \cite{tong2022videomae}}                        & \multicolumn{1}{c|}{ViT-B}                               & \multicolumn{1}{c|}{}                                    & \multicolumn{1}{c|}{16 $\times$ 3 $\times$ 5}         & \multicolumn{1}{c|}{87}                                                                     & \multicolumn{1}{c|}{87}                                                                                              & \multicolumn{1}{c|}{2.7}                                                                            & 81.5            & 95.1            \\
\multicolumn{1}{l|}{VideoMAE-L \cite{tong2022videomae}}                        & \multicolumn{1}{c|}{ViT-L}                               & \multicolumn{1}{c|}{}                                    & \multicolumn{1}{c|}{16 $\times$ 3 $\times$ 5}         & \multicolumn{1}{c|}{305}                                                                    & \multicolumn{1}{c|}{305}                                                                                             & \multicolumn{1}{c|}{9.0}                                                                            & 85.2            & 96.8            \\
\multicolumn{1}{l|}{VideoMAE-L \cite{tong2022videomae}}                        & \multicolumn{1}{c|}{ViT-L}                               & \multicolumn{1}{c|}{}                                    & \multicolumn{1}{c|}{40 $\times$ 3 $\times$ 4}         & \multicolumn{1}{c|}{305}                                                                    & \multicolumn{1}{c|}{305}                                                                                             & \multicolumn{1}{c|}{47.5}                                                                           & 86.1            & 97.3            \\ \hline
\multicolumn{9}{l}{\textit{Well-prepared ViT with plug-and-play modules.}}                                                                                                                                                                                                                                                                                                                                                                                                                                                                                                                          \\
\multicolumn{1}{l|}{TimeSformer-L \cite{bertasius2021space}}                     & \multicolumn{1}{c|}{ViT-B}                               & \multicolumn{1}{c|}{IN-21K}                              & \multicolumn{1}{c|}{96 $\times$ 3 $\times$ 1}         & \multicolumn{1}{c|}{121}                                                                    & \multicolumn{1}{c|}{121}                                                                                             & \multicolumn{1}{c|}{7.1}                                                                            & 80.7            & 94.7            \\
\multicolumn{1}{l|}{CoCa \cite{yu2022coca}}                              & \multicolumn{1}{c|}{ViT-g}                               & \multicolumn{1}{c|}{JFT-3B+ALIGN-1.8B}                   & \multicolumn{1}{c|}{N/A}                              & \multicolumn{1}{c|}{1000+}                                                                  & \multicolumn{1}{c|}{1000+}                                                                                           & \multicolumn{1}{c|}{N/A}                                                                            & \textbf{88.9}            & -               \\
\multicolumn{1}{l|}{MTV-H \cite{yan2022multiview}}                             & \multicolumn{1}{c|}{ViT-H+B+S+T}                         & \multicolumn{1}{c|}{IN-21K+WTS-600M}                     & \multicolumn{1}{c|}{32 $\times$ 3 $\times$ 4}         & \multicolumn{1}{c|}{1000+}                                                                  & \multicolumn{1}{c|}{1000+}                                                                                           & \multicolumn{1}{c|}{44.5}                                                                           & \textbf{89.1}            & 98.2            \\ \hline
\multicolumn{9}{l}{\textit{Full tuning}}                                                                                                                                                                                                                                                                                                                                                                                                                                                                                                                                                            \\
\multicolumn{1}{l|}{UniFormerV2-B \cite{li2023uniformerv2}}                     & \multicolumn{1}{c|}{ViT-B}                               & \multicolumn{1}{c|}{CLIP-400M}                           & \multicolumn{1}{c|}{8 $\times$ 3 $\times$ 4}          & \multicolumn{1}{c|}{115}                                                                    & \multicolumn{1}{c|}{115}                                                                                             & \multicolumn{1}{c|}{1.6}                                                                            & 84.4            & 96.3            \\
\multicolumn{1}{l|}{UniFormerV2-L \cite{li2023uniformerv2}}                     & \multicolumn{1}{c|}{ViT-L}                               & \multicolumn{1}{c|}{CLIP-400M}                           & \multicolumn{1}{c|}{8 $\times$ 3 $\times$ 4}          & \multicolumn{1}{c|}{354}                                                                    & \multicolumn{1}{c|}{354}                                                                                             & \multicolumn{1}{c|}{8.0}                                                                            & 87.7            & 97.9            \\
\multicolumn{1}{l|}{UniFormerV2-B \cite{li2023uniformerv2}}                     & \multicolumn{1}{c|}{ViT-B}                               & \multicolumn{1}{c|}{CLIP-400M+K710-0.66M}                & \multicolumn{1}{c|}{8 $\times$ 3 $\times$ 4}          & \multicolumn{1}{c|}{115}                                                                    & \multicolumn{1}{c|}{115}                                                                                             & \multicolumn{1}{c|}{1.6}                                                                            & 85.6            & 97.0            \\
\multicolumn{1}{l|}{UniFormerV2-L \cite{li2023uniformerv2}}                     & \multicolumn{1}{c|}{ViT-L}                               & \multicolumn{1}{c|}{CLIP-400M+K710-0.66M}                & \multicolumn{1}{c|}{32 $\times$ 3 $\times$ 4}         & \multicolumn{1}{c|}{354}                                                                    & \multicolumn{1}{c|}{354}                                                                                             & \multicolumn{1}{c|}{16.0}                                                                           & \textbf{89.7}            & 98.3            \\ \hline
\multicolumn{9}{l}{\textit{Parameter Efficient Tuning}}                                                                                                                                                                                                                                                                                                                                                                                                                                                                                                                                             \\
\multicolumn{1}{l|}{ST-Adapter \cite{stadapter}}                        & \multicolumn{1}{c|}{ViT-B}                               & \multicolumn{1}{c|}{CLIP-400M}                           & \multicolumn{1}{c|}{32 $\times$ 3 $\times$ 1}         & \multicolumn{1}{c|}{93}                                                                     & \multicolumn{1}{c|}{7}                                                                                               & \multicolumn{1}{c|}{1.8}                                                                            & 82.7            & 96.2            \\
\multicolumn{1}{l|}{DUALPATH-B \cite{dualpath}}                        & \multicolumn{1}{c|}{ViT-B}                               & \multicolumn{1}{c|}{CLIP-400M}                           & \multicolumn{1}{c|}{32 $\times$ 3 $\times$ 1}         & \multicolumn{1}{c|}{96}                                                                     & \multicolumn{1}{c|}{10}                                                                                              & \multicolumn{1}{c|}{0.7}                                                                            & 85.4            & 97.1            \\
\multicolumn{1}{l|}{DUALPATH-L \cite{dualpath}}                        & \multicolumn{1}{c|}{ViT-L}                               & \multicolumn{1}{c|}{CLIP-400M}                           & \multicolumn{1}{c|}{32 $\times$ 3 $\times$ 1}         & \multicolumn{1}{c|}{330}                                                                    & \multicolumn{1}{c|}{27}                                                                                              & \multicolumn{1}{c|}{1.9}                                                                            & 87.7            & 97.8            
\\ \hline
\multicolumn{1}{l|}{{Ours - AM/12, LSTM}}                     & \multicolumn{1}{c|}{{ViT-B(Dinov2)}}                      & \multicolumn{1}{c|}{{IN-21K}}                   & \multicolumn{1}{c|}{{8 $\times$ 3 $\times$ 1}} & \multicolumn{1}{c|}{{86+32}+45+360}                                                              & \multicolumn{1}{c|}{32+45+360}                                                                                       & \multicolumn{1}{c|}{{5.3}}                                                                      & {88.8}       & {98.2}       \\ 
\multicolumn{1}{l|}{{Ours - AM/12, Transformer}}                     & \multicolumn{1}{c|}{{ViT-B(Dinov2)}}                      & \multicolumn{1}{c|}{{IN-21K}}                   & \multicolumn{1}{c|}{{8 $\times$ 3 $\times$ 1}} & \multicolumn{1}{c|}{{86+32}+45+103}                                                              & \multicolumn{1}{c|}{{32}+45+103}                                                                                       & \multicolumn{1}{c|}{{6.9}}                                                                      & {89.1}       & {98.3}       \\
\multicolumn{1}{l|}{\textbf{Ours - AM/12, Transformer}}                     & \multicolumn{1}{c|}{\textbf{ViT-B(Dinov2)}}                      & \multicolumn{1}{c|}{\textbf{IN-21K}}                   & \multicolumn{1}{c|}{\textbf{32 $\times$ 3 $\times$ 1}} & \multicolumn{1}{c|}{\textbf{86+32+45+103}}                                                              & \multicolumn{1}{c|}{\textbf{32+45+103}}                                                                                       & \multicolumn{1}{c|}{\textbf{13.8}}                                                                      & \textbf{89.6}       & \textbf{98.4}       \\

\hline
\end{tabular}
}
\caption{Comparison for K400 with the SOTA. The parameters are broken into components: backbone, {AM flow and linear layers}, {aligning encoder}, and {temporal processing module}. AM/12 signifies that AM flow is added to all 12 transformer blocks.} \label{K400}
\end{table*}


 
 

\subsection{Specific Details and Comparison to SOTA}
\label{4.2}
In this section, the performance of our model is compared against the baselines and the respective SOTA \shortcite{yan2022multiview, li2023uniformerv2, tong2022videomae, das2021vpn++} for the datasets. Training details are given in the appendix.
\subsubsection{SSv2}
We achieve the SOTA results for Ssv2 as in Table \ref{ssv2}.
Since this is a more challenging dataset and 8 frames are not enough to capture the essence of the actions, we also experiment with 24 frames as input and use the computation of AM flow with the aligning encoder. Results for both settings are presented. 
We also show results using CLIP backbone in place of dinov2 and these are discussed in the ablation study.

Adapter and TPM are added to each transformer block as SSv2 is challenging and very far from the pretraining data distribution, so it requires modification for each block.

\subsubsection{K400}
We report the SOTA comparison for K400 in Table \ref{K400}.
We achieve the best results with 32 frames and using a transformer encoder for temporal processing (discussed on page \pageref{ablation}). But we also achieve high performance when using only 8 frames and pretraining just on ImageNet dataset. We lag behind Uniformerv2-L, but compared to them, we have negligible pretraining data and require fewer frames (8 for us vs.\ 32), training time (160 GPU hours for us vs.\ 9600), flops (5.3T for us vs.\ 16T), and backbone (ViT-B for us vs.\ ViT-L) to achieve comparable results. Comparison against CoCa and MTV-H is not fair, as they have 1B+ parameters.
Our number of parameters is high, as the aligning encoder and the TPMs are not optimised for their function and are only used as a proof of concept. Therefore, our total number of parameters is reducible but is left for future work (discussed further on page \pageref{ablation}).

The aligning encoder and AM flow computation are employed since the camera is moving. Each transformer block has temporal processing modules and adapters attached to it.

\begin{table}[h]
\begin{center}
\resizebox{\columnwidth}{!}{
\begin{tabular}{c|cc|c|l|c}
\hline
Method                    & \multicolumn{1}{c|}{RGB}    & Skeleton                         & Pretrain              & \# F & \textit{CS} \\ \hline
Separable STA \cite{climent2021improved}   64          & \checkmark & \checkmark & K400                  &   -   & 54.2                         \\
\vspace{0.7em} VPN \cite{das2020vpn}                      & \checkmark & \checkmark & K400                  &   64 +pose   & 60.8                         \\
\vspace{0.7em} MMNet \cite{bruce2022mmnet}                     & \checkmark & \checkmark & -                     &    -  & 70.1                         \\
\vspace{0.7em} VPN++ \cite{das2021vpn++}           & \checkmark & \checkmark & -                     &  64 +pose   & 69.0                         \\ \hline
ST-GCN \cite{yan2018spatial}&                           & \checkmark & Scratch               &  -    & 53.8                         \\
\vspace{0.7em} 2s-AGCN \cite{shi2019two}                  &                           & \checkmark & Scratch               &   -   & 60.9                         \\
\vspace{0.7em} MS-G3D \cite{liu2020disentangling}                    &                           & \checkmark & Scratch               &    -  & 61.1                         \\
\vspace{0.7em} UNIK \cite{yang2021unik}                      &                           & \checkmark & Posetics              &  -    & 64.3                         \\ \hline
I3D \cite{li2012cross}                       & \checkmark &                           & K400                 &  64  & 53.4                         \\
\vspace{0.7em} AssembleNet++\cite{ryoo2020assemblenet}              & \checkmark &                           & K400          &   -        & 63.6                         \\ \hline

\multicolumn{1}{c|}{\textbf{Ours - AM/2(1,12), LSTM}} & \checkmark &  \multicolumn{1}{l|}{}     & \multicolumn{1}{l|}{\textbf{IN-21K}} & \textbf{8}    & \multicolumn{1}{l}{\textbf{70.2}}         \\ \hline
\end{tabular}
}
\end{center}
\caption{Results of Toyota Smarthome. AM/2 signifies that AM flow is added to two transformer blocks (1,12). Thus, it has 1/6th the number of parameters for temporal and aligning encoders compared to kinetics-400 and SSv2.}
\label{smarthome}
\end{table}

\subsubsection{Toyota Smarthome}
The comparison for Smarthome is reported in Table \ref{smarthome} achieving SOTA results. This shows that our method also adapts well to small datasets.

The camera is fixed here, and since these are daily-action videos for older-adults, the subjects do not move much in the video. Therefore, the vanilla AM flow (without aligning encoder) is used. AM flow is added to the first and last transformer blocks.

\begin{table}[h]
\begin{center}
\resizebox{0.52\columnwidth}{!}{%
\begin{tabular}{cc}
\hline
\textbf{Model}                 & \textbf{Top-1} \\ \hline
w/o AM Flow           & 74.3  \\
w/o Aligning Encoder  & 72.7  \\
Training from scratch & 78.3 \\
Ours with hyperformer & 84.2  \\
Ours                  &  88.8     \\ \hline
\end{tabular}
}
\caption{Ablation study for Component Analysis (Experiments performed on K400).}
\label{tab:table1}
\end{center}
\end{table}

\subsection{Additional Experiments and Discussion}

\label{ablation}
This section covers ablation studies and other experiments to validate the efficacy of the contributions.

\textbf{Number of parameters and changing the temporal processing module and the input frames.}
The choice of temporal processing module is not focused on in this work as even with LSTMs, we get enough performance to prove the efficacy of the additions. TCN and transformer encoder provide the same performance with fewer parameters. The additions also do not have to be done at each step, and this further reduces the number of trainable parameters considerably. This is discussed in the following part of this subsection. But irrespective of the module used, as discussed above, we require very less training and have negligible pretraining requirements compared to the SOTA.

LSTMs do not have enough processing power to handle more than 8 frames as input. Since SSv2 has complex temporal relations, we use TCN and transformer encoder with 24 frames to achieve better performance. 
Using more number of frames for K400 also improves performance showing that the model is scalable.

\begin{figure}[ht]
\begin{center}
\scalebox{0.8}{
   \includegraphics[width=\linewidth]{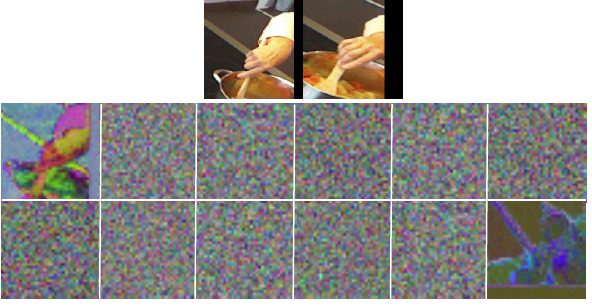}
   }
\end{center}
   \caption{Computed AM flow for two frames (on top) from Smarthome. Starting from top-left, going row-wise, AM flow is visualised for each transformer block in ViT-B from beginning to the end. The figure shows that we do not need to add AM flow to each layer and here for example, only the first and last layer are important.} \label{fig5}
\end{figure}

\textbf{Location of AM Flow addition.}
 The experiment on the Smarthome dataset is carried out to answer this, Table \ref{smarthome}. Adapters with AM flow and temporal processing modules are only added to the first and last blocks of the transformer. Traditional adapters are added to the rest of the blocks. We optimised the addition of these modules only for this dataset, to demonstrate that it is possible. The method to do this was to visualise AM flow by taking PCA along the channel dimension to reduce it to 3 dimensions (corresponding to RGB). Figure \ref{fig5} clarifies that the first and the last blocks produce visualisations close to what AM flow is supposed to signify, as discussed in previous sections.

 For larger datasets such as K400 and SSv2, we need to add AM flow and the temporal processing module to more number of backbone blocks, but for a small dataset like smarthome, a small number of blocks are enough and thus the trainable parameters are reduced. This alleviates the challenge of having a large model and fewer training data.

\textbf{Impact of AM Flow.}
Results in Table \ref{tab:table1}. We observe that the addition of AM flow greatly improves performance. Two factors affect this improvement:
1) Concatenation of AM flow to the adapter input pushes the downsampled embedding to learn semantic information about the input, thus parallel adapters overcome their shortcoming over serial adapters for temporal processing (details in the appendix). 
2) As intended, it adds local temporal attention to the model.

\textbf{Impact of aligning encoder.}
Results in Table \ref{tab:table1}. With irrelevant motion in the frames, AM flow is noisy, as the patches with change are not only due to the action but also to motion. The aligning encoder resolves this issue (as demonstrated in the appendix).

\textbf{Changing type of adapter.}
Results in Table \ref{tab:table1}. There are various variations of adapters, such as hyperformer \cite{karimi2021parameter} and compacter \cite{karimi2021compacter}. We compare against hyperformer and observe slightly poor performance, but hyperformer is also more parameter efficient than a simple adapter. The choice to use either is to play with this tradeoff.

\textbf{Training from scratch.} To make sure that a well-chosen pretrained image model is important for our work, we train from scratch with the same architecture and achieve 78.3\% (as in Table \ref{tab:table1}) accuracy for K400 as compared to 88.8\% in Table \ref{K400}. This shows our method succeeds in utilising the spatial awareness learnt by pretrained image models and it is an important step.

\textbf{Changing the backbone.} For completeness, we use CLIP as the backbone in place of dinov2 in Table \ref{ssv2} and show that we still achieve SOTA results and that our method is not specific to the pretraining method. We show results with Hiera \shortcite{ryali2023hiera} in the appendix. It is important to have good spatial features from the backbone, irrespective of its nature.

 \section{Conclusion and Future Work}
In this paper, we have introduced a novel image-to-video transfer learning method based on two ideas.
Firstly, we compute motion information from the attention maps of the image backbone. We do this using the newly introduced concept of AM flow. We also provide two formulations of AM flow depending on camera movement. Secondly, using temporal processing adapters we add AM flow into a frozen pretrained image model and utilise the downsampled embedding obtained from the adapters for global temporal processing. 
We presented experiments on large datasets (K400, SSv2) with a low number of training steps required for convergence and using only ImageNet for pretraining. The model is additionally able to adapt to small datasets, shown by our experiments on the Toyota Smarthome dataset where we achieve SOTA results too.
Through ablation studies, we verify the pertinence of various parts of the model. We achieve SOTA or comparable performance for the three chosen datasets.
We obtain this performance with only the ImageNet dataset for pretraining, while reducing training time.

There are various avenues to extend this work. Adding memory to AM flow should lead to more nuanced temporal information. The work can thus be extended to detection tasks in videos. 
Further, as the aligning encoder is resource intensive, we intend to explore 
machine learning alternatives in future work. 

\section*{Ethical Statement}
This research adheres to the highest ethical standards, ensuring the welfare, dignity, and rights of all individuals involved. Even though this work does not record new data, the used data was collected with an informed consent obtained from all participants. No harm or bias has been detected during our research activities. We contribute to the advancement of knowledge while prioritizing the well-being of those involved within the scope of the European GDPR regulations.

\bibliography{aaai25}

\end{document}